# Effect of Input Resolution on Retinal Vessel Segmentation Performance: An Empirical Study Across Five Datasets

Amarnath R
*United Institute of Technology, Coimbatore, India*
**Date:** April 3, 2026
**Correspondence:** amarnathresearch@gmail.com

**Abstract**

Most deep learning pipelines for retinal vessel segmentation resize fundus images to satisfy GPU memory constraints and enable uniform batch processing. However, the impact of this resizing on thin vessel detection remains underexplored. When high resolution images are downsampled, thin vessels are reduced to subpixel structures, causing irreversible information loss even before the data enters the network. Standard volumetric metrics such as the Dice score do not capture this loss because thick vessel pixels dominate the evaluation. We investigated this effect by training a baseline UNet at multiple downsampling ratios across five fundus datasets (DRIVE, STARE, CHASE_DB1, HRF, and FIVES) with native widths ranging from 565 to 3504 pixels, keeping all other settings fixed. We introduce a width-stratified sensitivity metric that evaluates thin (half-width <3 pixels), medium (3 to 7 pixels), and thick (>7 pixels) vessel detection separately, using native resolution width estimates derived from a Euclidean distance transform. Results show that for high-resolution datasets (HRF, FIVES), thin vessel sensitivity improves monotonically as images are downsampled toward the encoder's effective operating range, peaking at processed widths between 256 and 876 pixels. For low-to-mid resolution datasets (DRIVE, STARE, CHASE_DB1), thin vessel sensitivity is highest at or near native resolution and degrades with any downsampling. Across all five datasets, aggressive downsampling reduced thin vessel sensitivity by up to 15.8 percentage points (DRIVE) while Dice remained relatively stable, confirming that Dice alone is insufficient for evaluating microvascular segmentation.

**Keywords:** retinal vessel segmentation, input resolution, thin vessel detection, encoder-decoder networks, width-stratified sensitivity, fundus imaging

## Highlights

- Input resolution effects tested independently across five fundus datasets spanning a 6x width range (565 to 3504 pixels).
- New width-stratified metric reveals thin vessel losses hidden by Dice score.
- High-resolution datasets (HRF, FIVES) show monotonic thin vessel sensitivity improvement with downsampling; low-resolution datasets peak at native resolution.
- Aggressive downsampling reduces thin vessel sensitivity by up to 15.8pp while Dice remains stable.
- Standard pooling networks are inherently constrained for high resolution clinical imaging.

## 1. Introduction

The accurate detection of thin peripheral vessels in fundus images is critical for automated screening, as early microvascular alterations including capillary dropout, microaneurysms, and peripheral tortuosity changes are primary indicators of diabetic and hypertensive retinopathy [1, 2]. These alterations occur exclusively in the thinnest vessel branches, structures that are often only one to two pixels wide even at native acquisition resolution. However, public datasets vary significantly in native resolution (e.g., 565 × 584 pixels in DRIVE to 3504 × 2336 pixels in HRF) because clinical images are acquired at high resolutions for detailed investigation. To maintain the structural layout of the retina while satisfying GPU memory constraints and enabling uniform batch processing, most deep learning pipelines downsample whole images to a fixed target size. Consequently, at large reduction ratios, thin vessels are compressed into subpixel structures, effectively blurring or erasing microvascular details and causing structural information loss before network processing [4, 8]. Standard evaluation metrics fail to show this loss. Specifically, the Dice score is dominated by larger, thicker vessels, allowing models to report high overall performance even if they completely miss thin vessels.

While resolution variance has been noted as a factor in cross dataset generalization [4, 8], the specific structural loss incurred by input decimation has not been systematically quantified in retinal vessel segmentation. The practical consequence is that published Dice scores across datasets are not equivalent. A model evaluated on DRIVE at near native resolution and a model evaluated on HRF at 6.9x downsampling operate under substantially different thin vessel conditions, yet both report a single Dice value. This makes cross dataset comparisons unreliable and obscures whether reported improvements reflect genuine model advances or resolution artifacts. Quantifying these resizing losses is therefore essential to interpret evaluation metrics accurately and improve the reliability of automated screening. While patch-based training at native resolution can bypass GPU memory constraints entirely, most published retinal vessel segmentation benchmarks employ whole image resizing pipelines [3, 4, 8], making the quantification of resizing induced structural loss directly relevant to the interpretation of reported results in the literature.

In this paper, we present a controlled empirical study testing input resolution as the primary experimental variable across five public fundus datasets. We use a standard UNet to ensure that changes in performance are caused only by resolution, not by complex model features. Our core contributions are threefold: (1) a systematic test across five datasets using the same settings to isolate the specific effect of image resizing; (2) a width-stratified sensitivity metric that evaluates thin, medium, and thick vessels separately using a distance transform based width estimate at native resolution; and (3) a precise measurement of the performance change at each downsampling ratio, showing that the same ratio produces opposite effects on HRF and DRIVE.

## 2. Related Work

Retinal vessel segmentation has been studied extensively across architectures, datasets, and loss functions [3, 7], however, input resolution as an independent variable has never been empirically examined. Galdran et al. [4] observed that a model trained on DRIVE failed to generalize to HRF and attributed this to the large resolution gap but did not systematically vary the resizing ratio. Similarly, Fadugba et al. [8] identified image quality as the primary factor in cross dataset performance without isolating resolution. Neither study measured the effect of downsampling on thin vessel detection.

Recent advancements in retinal vessel segmentation have increasingly focused on complex architectural solutions to improve fine vessel detection, such as multi featured approaches [5] and detail enhanced attention networks [6]. However, these architectural innovations do not address the foundational structural loss incurred during standard input decimation.

Multiscale architectures attempt to recover the fine vessel detail discarded by standard downsampling. Feature pyramid networks [9] aggregate features across decoder levels, while parallel branch designs such as NFN+ [10] and SCS-Net [12] process images at multiple input sizes simultaneously. Whether the optimal scale contribution is learned or fixed, the choice implicitly reflects the resolution range at which the encoder operates most effectively. These methods address information loss by increasing architectural complexity. The present study instead quantifies the resolution constraint directly through single-scale ablation, isolating it from other architectural variables.

Quantifying this effect is impossible with standard evaluation metrics, which inherently obscure microvascular performance. The RETA benchmark [13] incorporates topology metrics but does not report sensitivity by vessel width. Recent work has addressed volumetric bias in tubular structure segmentation through topological loss functions, including centerline Dice (clDice) [20], Skeleton Distance Loss [20], and Betti matching approaches [21, 22], which reweight thin structure gradients during training. The width-stratified sensitivity metric proposed here serves a complementary but distinct purpose. Rather than modifying the training objective, it provides a resolution independent auditing tool to make visible the structural losses that occur before network processing, and ultimately, losses that topological information cannot recover once subpixel aliasing has already occurred during preprocessing. No published study on retinal segmentation evaluates thin, medium, and thick vessel detection independently as a function of input resolution. The width-stratified metric introduced here provides the granularity needed to fill this gap.

## 3. Datasets and Experimental Setup

### 3.1. Datasets

Five publicly available fundus datasets were used in this study: DRIVE [11], STARE [14], CHASE_DB1 [15], HRF [16], and FIVES [17]. Their properties are summarized in Table 1. Native image widths span a 6x range, from 565 pixels to 3504 pixels, covering low resolution, mid resolution, and high resolution acquisition conditions. All images are 8 bit RGB with manually annotated binary vessel masks.

For DRIVE, the 20 training set images were used under fivefold cross validation. The 20 test set images were excluded because they carry no publicly available second expert annotation, making consistent ground truth comparison across folds less reliable. Using the training partition exclusively ensures all folds use identically annotated ground truth from the same annotation protocol. For FIVES, a stratified subset of 25 images per disease category (AMD, Normal, DR, Glaucoma) was drawn from the full 800 image dataset to ensure disease balanced evaluation at a computationally feasible scale.

Table 1. Summary of the five publicly available fundus datasets used in this study. Native image widths span a 6x range from 565 pixels to 3504 pixels. N refers to images used in this study. For FIVES, 100 images were drawn from the full 800 image dataset using stratified sampling of 25 images per disease category.

| Dataset | N | Native Size | Camera | Population | Tested | Width Ratio |
|---|---|---|---|---|---|---|
| DRIVE | 20 | 565 x 584 | Canon CR5 non-mydriatic, 45 deg FOV | Adults aged 25-90, diabetic retinopathy screening, Netherlands | R1 to R5 | 1.0x |
| STARE | 20 | 700 x 605 | Topcon TRV-50, 35 deg FOV | Mixed healthy and high-pathology adults, UC San Diego | R1 to R5 | 1.2x |
| CHASE_DB1 | 28 | 1280 x 960 | Nidek NM-200-D handheld, 30 deg FOV | Multi-ethnic school children, mean age 10, London | R1 to R5 | 2.3x |
| HRF | 45 | 3504 x 2336 | Canon CR-1 mydriatic, 45 deg FOV | Adults: 15 healthy, 15 diabetic retinopathy, 15 glaucoma, Czech Republic | R1 to R5 | 6.2x |
| FIVES | 100 [800] | 2048 x 2048 | Not specified in dataset paper | Adults: 200 DR, 200 AMD, 200 glaucoma, 200 normal, Zhejiang, China | R1 to R5 | 3.6x |

### 3.2. Resizing Conditions

Five resizing conditions were defined per dataset. Table 2 lists all conditions and their corresponding processed sizes and downsampling ratios. Conditions were selected to cover the range from native resolution down to aggressively downsampled sizes. For DRIVE and STARE an additional upsampled condition at 512×512 was included to test whether bilinear upsampling improves performance on low resolution images. All images were resized with bilinear interpolation. All masks were resized with the nearest neighbor interpolation to preserve binary boundaries. These interpolation choices were held fixed across all conditions and datasets to ensure that any observed performance difference is attributable to the change in processed resolution rather than the resampling method.

Table 2. Processing Sizes and Downsampling Ratios

| Dataset | Condition | Processed Size | Ratio |
|---|---|---|---|
| DRIVE | R1 | 565 x 584 | 1.0x (native) |
| DRIVE | R2 | 423 x 438 | 1.3x |
| DRIVE | R3 | 282 x 292 | 2.0x |
| DRIVE | R4 | 141 x 146 | 4.0x |
| DRIVE | R5 | 512 x 512 | upsampled |
| STARE | R1 | 700 x 605 | 1.0x (native) |
| STARE | R2 | 525 x 453 | 1.3x |
| STARE | R3 | 350 x 302 | 2.0x |
| STARE | R4 | 175 x 151 | 4.0x |
| STARE | R5 | 512 x 512 | upsampled |
| CHASE_DB1 | R1 | 1280 x 960 | 1.0x (native) |
| CHASE_DB1 | R2 | 960 x 720 | 1.3x |
| CHASE_DB1 | R3 | 640 x 480 | 2.0x |
| CHASE_DB1 | R4 | 320 x 240 | 4.0x |
| CHASE_DB1 | R5 | 512 x 512 | standard |
| HRF | R1 | 3504×2336 | 1.0x (native) |
| HRF | R2 | 2628×1752 | 1.3x |
| HRF | R3 | 1752 x 1168 | 2.0x |
| HRF | R4 | 876 x 584 | 4.0x |
| HRF | R5 | 512 x 512 | 6.9x |
| FIVES | R1 | 2048×2048 | 1.0x (native) |
| FIVES | R2 | 1536×1536 | 1.3x |
| FIVES | R3 | 1024 x 1024 | 2.0x |
| FIVES | R4 | 512 x 512 | 4.0x |
| FIVES | R5 | 256 x 256 | 8.0x |

### 3.3. Model

A UNet [4] with four encoder stages and channel sizes [16, 32, 64, 128] (1.9M parameters) was used as the sole architecture for all resolution conditions across all datasets [18]. Each encoder stage consists of two convolutional blocks followed by 2×2 max pooling. The decoder mirrors the encoder with bilinear upsampling and skip connections. The loss function combines Dice loss and binary cross entropy with equal weighting. This architecture was selected because Galdran et al. [4] demonstrated that a minimal UNet achieves performance close to state-of-the-art on these datasets, making it a suitable controlled baseline for isolating input resolution as the sole experimental variable. To fit the largest inputs (HRF at 3504×2336) within the 16 GB memory budget of a T4 GPU, mixed-precision training (float16), gradient checkpointing on all encoder blocks and the bottleneck, and gradient accumulation over 4 steps (effective batch size 4) were used. All other settings were held fixed across all conditions and datasets: AdamW optimizer [19], learning rate $1\times10^{-3}$, weight decay $1\times10^{-4}$, batch size 1 with 4-step gradient accumulation, maximum 300 epochs, early stopping at patience 15 on validation Dice, and random seed 42.

### 3.4. Width-Stratified Sensitivity

Standard sensitivity treats all vessel pixels equally regardless of vessel width. Because thick vessels contribute more pixels to the total count, a model that fails to detect thin vessels can still achieve high overall sensitivity. To make the effect of resolution on fine vascular structures visible, a width-stratified sensitivity metric was computed as follows.

A Euclidean distance transform was applied to each ground truth mask at native resolution. The distance value at each vessel pixel provides an estimate of the local vessel half width. Vessel pixels were grouped into three categories such as thin (half width below 3 pixels), medium (3 to 7 pixels), and thick (above 7 pixels). Both the predicted segmentation and the ground truth mask were upsampled to native resolution before this calculation to ensure width estimates are consistent across all resolution conditions. For predictions generated at reduced resolution, this upsampling step may introduce jagged boundaries that slightly alter apparent vessel width at native scale. This artefact is uniform across all conditions using the same interpolation method and does not affect within-dataset relative comparisons. Sensitivity was computed independently within each group as the ratio of correctly detected vessel pixels to total ground truth vessel pixels in that group, averaged across images and then across folds.

Formally, let $W_k$ denote the set of vessel pixels in width stratum k, where k is thin, medium, or thick, defined by half width thresholds of 3 pixels and 7 pixels at native resolution. Width-stratified sensitivity for stratum k is computed as:

$$S_k = \frac{\sum_{i \in W_k} 1[y'_i = 1, y_i = 1]}{\sum_{i \in W_k} 1[y_i = 1]} \tag{1}$$

where $y_i$ is the ground truth label and $y'_i$ is the predicted label after upsampling both maps to native resolution. Standard sensitivity corresponds to the case where $W_k$ covers all vessel pixels regardless of width. The relationship between Dice score and thin vessel sensitivity across all conditions is shown in Figure 1, demonstrating that high Dice does not imply adequate thin vessel detection.

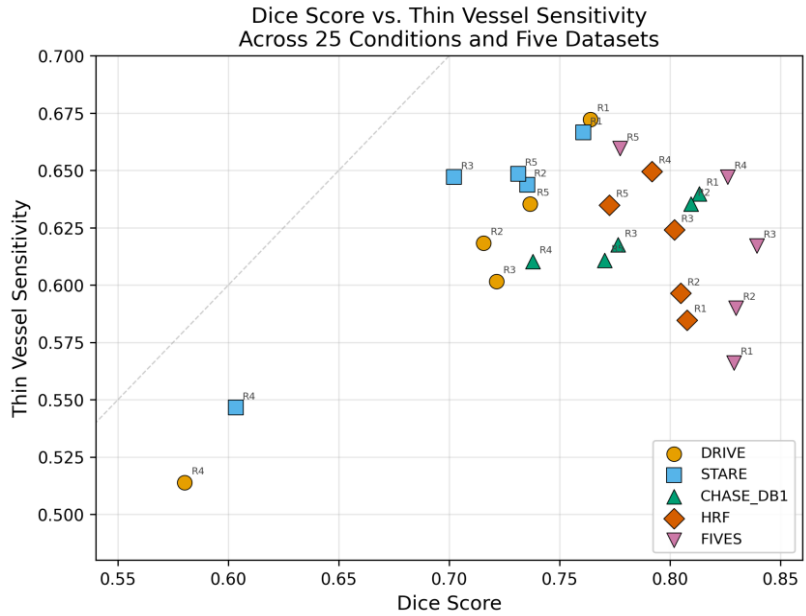

**Fig. 1** Dice score versus thin vessel sensitivity across all 25 conditions and five datasets. Each point represents one dataset-condition pair. High Dice does not reliably predict high thin vessel sensitivity, demonstrating the volumetric bias of the Dice metric.

### 3.5. Evaluation

Fivefold cross validation with stratified fold assignments was used across all datasets. This design was chosen over official train/test splits to ensure balanced and comparable resolution evaluations across the small dataset sizes in this study. All predictions use a threshold of 0.5 on the sigmoid output. Mean and standard deviation across folds is reported for Dice, overall sensitivity, specificity, and thin vessel sensitivity. Full results are reported in Table 3 and thin vessel sensitivity across all processed widths is illustrated in Figure 2.

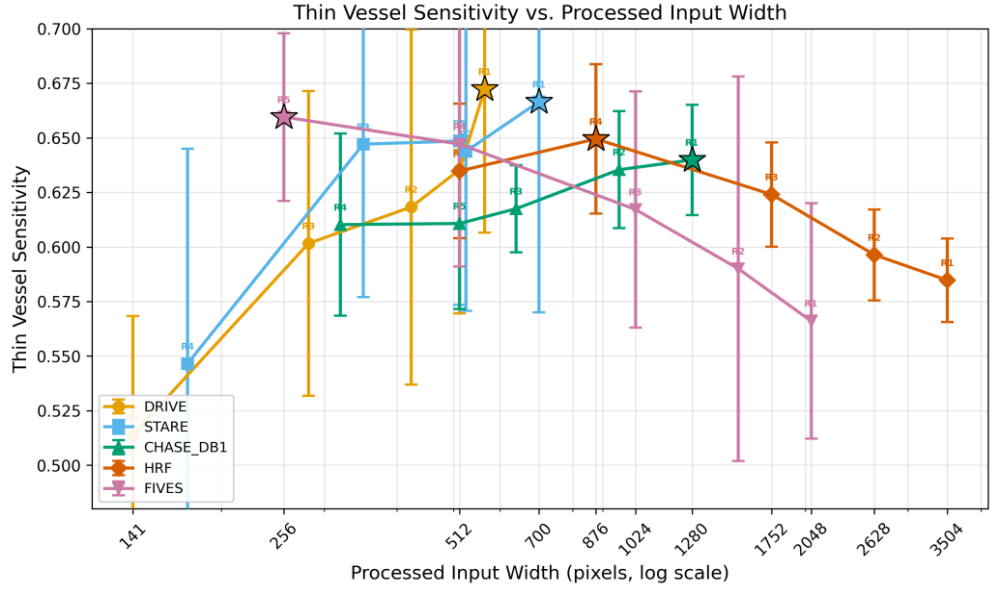

**Fig. 2** Thin vessel sensitivity as a function of processed input width across all five datasets. High-resolution datasets (HRF, FIVES) show monotonic improvement with downsampling, while low-to-mid resolution datasets (DRIVE, STARE, CHASE_DB1) peak at or near native resolution.

Table 3. *Mean 5-fold results for all datasets and conditions. Best thin vessel sensitivity per dataset in bold.*

| Dataset | Cond | Width | Dice | Sens | Spec | Thin Sens |
|---|---|---|---|---|---|---|
| DRIVE | R1 | 565px | 0.7639 | 0.7682 | 0.9778 | **0.6723** |
| DRIVE | R2 | 423px | 0.7156 | 0.7153 | 0.9753 | 0.6183 |
| DRIVE | R3 | 282px | 0.7214 | 0.6911 | 0.9802 | 0.6016 |
| DRIVE | R4 | 141px | 0.5803 | 0.5748 | 0.9643 | 0.5139 |
| DRIVE | R5 | 512px | 0.7366 | 0.7319 | 0.978 | 0.6354 |
| STARE | R1 | 700px | 0.7608 | 0.758 | 0.9817 | **0.6665** |
| STARE | R2 | 525px | 0.7353 | 0.7354 | 0.9799 | 0.6438 |
| STARE | R3 | 350px | 0.7022 | 0.7203 | 0.9746 | 0.6471 |
| STARE | R4 | 175px | 0.6034 | 0.5971 | 0.9713 | 0.5465 |
| STARE | R5 | 512px | 0.7312 | 0.7384 | 0.9772 | 0.6485 |
| CHASE_DB1 | R1 | 1280px | 0.8133 | 0.8245 | 0.9852 | **0.6399** |
| CHASE_DB1 | R2 | 960px | 0.8095 | 0.8136 | 0.9857 | 0.6354 |
| CHASE_DB1 | R3 | 640px | 0.7764 | 0.7746 | 0.984 | 0.6175 |
| CHASE_DB1 | R4 | 320px | 0.7379 | 0.7257 | 0.9824 | 0.6102 |
| CHASE_DB1 | R5 | 512px | 0.7704 | 0.7675 | 0.9838 | 0.6107 |
| HRF | R1 | 3504px | 0.8078 | 0.8211 | 0.9832 | 0.5847 |
| HRF | R2 | 2628px | 0.8048 | 0.8171 | 0.9831 | 0.5964 |
| HRF | R3 | 1752px | 0.8019 | 0.8204 | 0.9821 | 0.624 |
| HRF | R4 | 876px | 0.7918 | 0.7951 | 0.9831 | **0.6495** |
| HRF | R5 | 512px | 0.7725 | 0.7625 | 0.9832 | 0.6348 |
| FIVES | R1 | 2048px | 0.8289 | 0.7903 | 0.994 | 0.566 |
| FIVES | R2 | 1536px | 0.8299 | 0.798 | 0.9928 | 0.59 |
| FIVES | R3 | 1024px | 0.8394 | 0.8005 | 0.9945 | 0.6172 |
| FIVES | R4 | 512px | 0.8261 | 0.7808 | 0.9944 | 0.6471 |
| FIVES | R5 | 256px | 0.7773 | 0.7523 | 0.9873 | **0.6595** |

## 4. Results

### 4.1. Results by Condition

Table 3 reports the mean 5-fold results for all datasets and conditions. The best thin vessel sensitivity per dataset is shown in bold. Across all five datasets and 25 conditions, thin vessel sensitivity ranges from 0.5139 to 0.6723, while Dice ranges from 0.5803 to 0.8394. The divergence between these two metrics across conditions is illustrated in Figure 1, which shows that high Dice does not reliably predict high thin vessel sensitivity.

### 4.2. High-Resolution Datasets

For HRF and FIVES, where native resolutions far exceed the encoder's receptive field capacity, downsampling consistently improved thin vessel sensitivity, as shown in Table 3. On HRF, thin vessel sensitivity increased monotonically from 0.5847 at native resolution (3504px) to 0.6495 at 4.0x downsampling (876px), a gain of 6.5 percentage points. Dice remained stable across these conditions (0.8078 vs 0.7918), confirming that Dice alone does not capture this improvement. The spatial loss of thin vessel detail at different resolutions is shown in the patch comparisons in Figure 3.

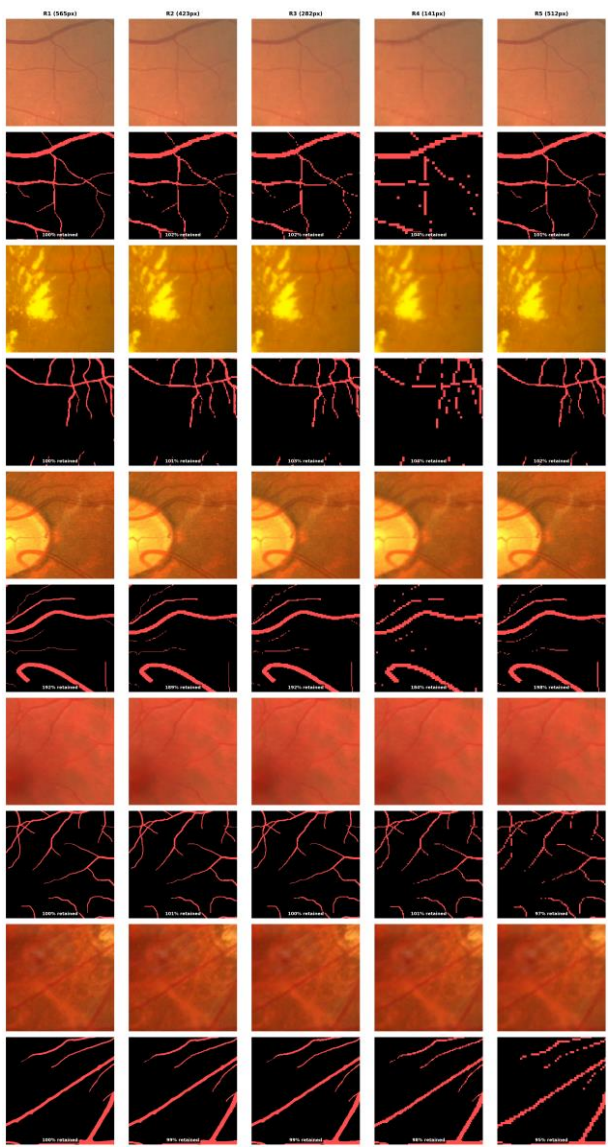

**Fig. 3** Patch comparisons showing thin vessel detail at different input resolutions of all datasets. In HRF, at 6.9x downsampling (512px) peripheral capillaries are erased before network processing

On FIVES, thin vessel sensitivity increased monotonically from 0.5660 at native resolution (2048px) to 0.6595 at 8.0x downsampling (256px), a gain of 9.4 percentage points. Dice remained above 0.82 for conditions R1 through R4 but dropped to 0.7773 at R5 (256px), indicating that the most aggressive downsampling begins to affect overall segmentation quality even as thin vessel detection continues to improve.

CHASE_DB1 showed a different pattern. Unlike HRF and FIVES, thin vessel sensitivity was highest at native resolution (0.6399 at 1280px) and declined monotonically with downsampling to 0.6102 at 4.0x (320px). However, the total range across conditions was narrow (3.0 percentage points), and the thin sensitivity curve was relatively flat compared to the other datasets. This suggests that CHASE_DB1 occupies a transitional position: its native resolution is high enough that the encoder can operate effectively without downsampling, but not so high that downsampling provides a benefit.

### 4.3. Low Resolution Datasets

For DRIVE and STARE, thin vessel sensitivity was highest at native resolution and declined with any downsampling, as shown in Table 3. On DRIVE, Dice fell from 0.7639 at native resolution (565px) to 0.5803 at 4.0x (141px), a drop of 18.4 percentage points. Thin vessel sensitivity fell from 0.6723 to 0.5139, a drop of 15.8 percentage points. On STARE, Dice fell from 0.7608 at native resolution (700px) to 0.6034 at 4.0x (175px), a drop of 15.7 percentage points. Thin vessel sensitivity fell from 0.6665 to 0.5465, a drop of 12.0 percentage points.

As visible in Figure 2, both DRIVE and STARE show a monotonic decline in thin vessel sensitivity as processed width decreases below native resolution. At 141px, DRIVE images are reduced to a scale where thick vessels occupy only 1 to 2 pixels and thin vessels fall below the pixel threshold entirely.

The upsampled condition R5 at 512×512 via bilinear interpolation partially recovered Dice on both DRIVE (0.7366 vs 0.7639) and STARE (0.7312 vs 0.7608), but thin vessel sensitivity remained lower than native in both cases (DRIVE: 0.6354 vs 0.6723, STARE: 0.6485 vs 0.6665). Bilinear upsampling does not recover vessel detail that was absent in the original image.

### 4.4. Effective Operating Range of the Encoder-Decoder

The best thin vessel sensitivity across all five datasets corresponds to processed widths of 565px (DRIVE, native), 700px (STARE, native), 1280px (CHASE_DB1, native), 876px (HRF, 4.0x), and 256px (FIVES, 8.0x). Three of these five optimal widths (DRIVE, STARE, HRF) fall within a 565 to 876 pixel range. However, CHASE_DB1 peaks at its native resolution of 1280px and FIVES peaks at 256px, indicating that no single processed width range is universally optimal.

The pattern is resolution-dependent rather than architecture-universal. For high-resolution datasets where native widths far exceed the encoder's receptive field (HRF at 3504px, FIVES at 2048px), downsampling consistently improves thin vessel sensitivity by bringing vessel widths into a range where the encoder can represent them as multi-pixel structures. For low-to-mid resolution datasets where native widths are closer to the encoder's operating range (DRIVE at 565px, STARE at 700px, CHASE_DB1 at 1280px), native resolution already provides the best thin vessel sensitivity and any downsampling degrades performance.

This dichotomy reflects the interaction between native vessel width and encoder capacity. A UNet with four max-pooling stages reduces spatial dimensions by a factor of 16 at the bottleneck. When the native image is large (e.g., HRF at 3504px), the bottleneck feature map at 219×146 pixels is too coarse relative to the thin vessel structures, and downsampling to 876px (bottleneck at 55×37) allows better context-to-detail balance. When the native image is small (e.g., DRIVE at 565px), the bottleneck is already at 35×37 pixels at native resolution, and further reduction pushes thin vessels below the pixel threshold entirely. The magnitude of the resolution effect across all five datasets is summarized in Figure 4, which shows the best and worst thin vessel sensitivity achieved per dataset across all tested conditions.

### 4.5. Statistical Significance

Statistical comparisons between resolution conditions were performed using Wilcoxon signed-rank tests on the five fold-level measurements. With n=5 paired observations, the minimum achievable two-sided p-value is 0.0625. We report exact p-values and adopt α=0.10 as the significance threshold. The difference between best and worst thin vessel sensitivity was marginally significant (p=0.0625) in four of five datasets (DRIVE, STARE, HRF, FIVES), with effect sizes ranging from 6.5 to 15.8 percentage points. CHASE_DB1 showed no significant difference (p=0.3125), consistent with its narrow thin vessel sensitivity range across conditions. Spearman correlation between Dice and thin vessel sensitivity across conditions was significantly negative for HRF (ρ=−0.90, p=0.037), indicating that Dice remained stable while thin vessel sensitivity improved with downsampling. This directly confirms that Dice alone does not capture resolution-dependent thin vessel performance on high-resolution datasets.

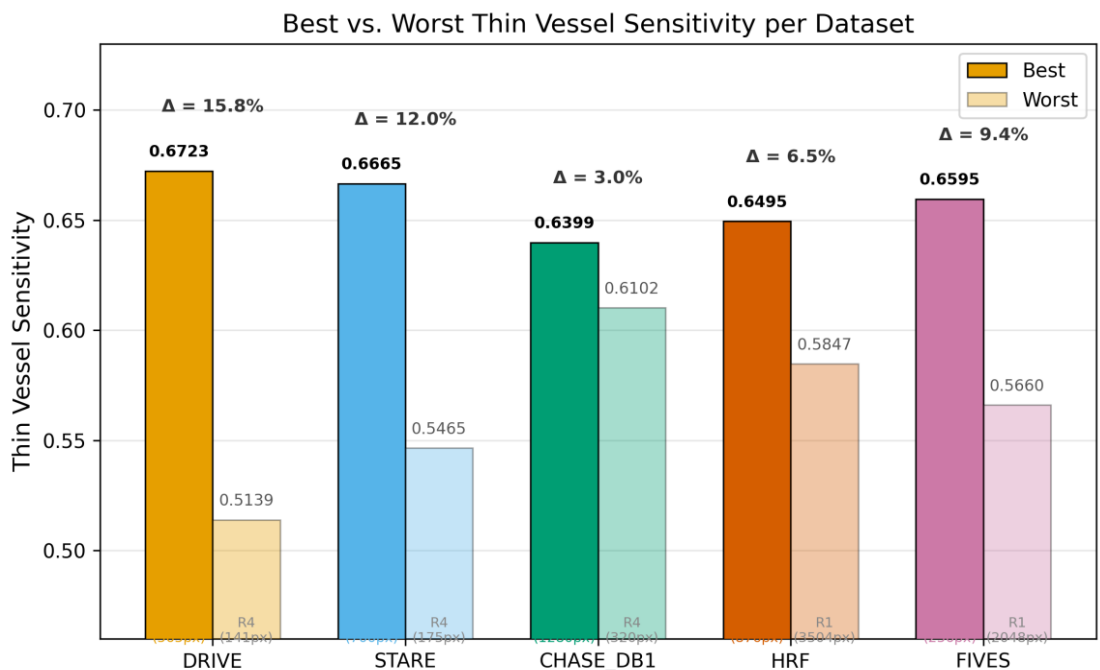

**Fig. 4** Best and worst thin vessel sensitivity per dataset across all tested resolution conditions. Solid bars show the highest thin vessel sensitivity achieved, and faded bars show the lowest. The gap ranges from 3.0 percentage points (CHASE_DB1) to 15.8 percentage points (DRIVE), confirming that input resolution choice has a substantial effect on thin vessel detection, with the magnitude depending on how far native resolution deviates from the encoder's effective operating range.

## 5. Discussion

### 5.1. Resolution Selection Relative to Native Image Size

The results reveal a resolution-dependent pattern in thin vessel sensitivity. For high-resolution datasets (HRF, FIVES), where native widths far exceed the encoder's receptive field capacity, downsampling consistently improves thin vessel detection. HRF thin vessel sensitivity improves from 0.5847 at native resolution to 0.6495 at 4.0x (876px), while FIVES improves from 0.5660 at native to 0.6595 at 8.0x (256px). For low-to-mid resolution datasets (DRIVE, STARE, CHASE_DB1), native resolution already provides optimal or near-optimal thin vessel sensitivity, and any downsampling degrades performance.

For HRF specifically, a processed width of 876 pixels (4.0x downsampling) outperforms 512×512 (6.9x downsampling) on thin vessel sensitivity (0.6495 vs 0.6348), suggesting that within the beneficial downsampling range, preserving more spatial detail is still advantageous and that the 512×512 convention used widely in the literature applies an unnecessarily aggressive ratio to high-resolution datasets. Conversely, on DRIVE and STARE, the native resolution is already near the encoder's effective operating range, and the common practice of processing these datasets at or near native resolution is empirically justified.

## 5.2. Benchmark Comparability

A practical consequence of this study concerns cross-dataset comparisons in the literature. Papers reporting results on DRIVE typically process images near native resolution (approximately 1.0x ratio). Papers reporting results on HRF frequently use 512×512 (approximately 6.9x ratio). The present results show these two processing choices produce substantially different thin vessel sensitivity patterns. A direct comparison of two methods where one is evaluated on DRIVE at native resolution and another on HRF at 6.9x downsampling is not methodologically equivalent, even if both report Dice scores.

The width-stratified sensitivity metric introduced here makes this inconsistency visible. Reporting this metric alongside Dice and overall sensitivity in multi dataset studies would allow more meaningful cross dataset comparisons.

## 5.3. Sensitivity-Precision Tradeoff at Very Small Input Sizes

On FIVES, the most aggressively downsampled condition R5 (256px) achieves the highest thin vessel sensitivity (0.6595) while Dice drops to its lowest value (0.7773). At 256×256 pixels the network assigns vessel labels to ambiguous low contrast regions that would be classified as background at larger input sizes. This increases thin vessel recall but also increases false positives. The appropriate resolution therefore depends on the clinical objective. In population level screening, maximising sensitivity to early microvascular change is the priority, and smaller input sizes within the effective range are preferable even at the cost of precision. In diagnostic confirmation settings, where segmentation accuracy and low false positive rates are critical, larger input sizes within the effective range are more appropriate. This tradeoff should be treated as a tunable parameter rather than a fixed convention, and the width-stratified sensitivity metric provides the granularity needed to make this choice visible and reproducible across studies.

## 5.4. Limitations

This study intentionally evaluated a single encoder-decoder architecture with four pooling stages to isolate input resolution as the sole experimental variable. Similarly, patch-based training at native resolution, which bypasses whole image resizing entirely, was excluded from this study by design. Quantifying the thin vessel sensitivity gap between optimally resized whole image pipelines and patch-based native resolution training is a natural and necessary extension of this work. The effective operating range identified here is specific to this architectural configuration, and whether analogous constraints exist for architectures with larger receptive fields, dilated convolutions, vision transformers, or full resolution pathways is an open question that this study motivates but does not address. Extending this study to such architecture is a natural direction for future work.

clDice and other topology-aware evaluation metrics were not computed in this study, as the focus was on isolating the effect of input resolution using a standard training pipeline without topology-preserving losses. Combining width-stratified evaluation with topological metrics across resolution conditions is left for future work.

The FIVES results are based on a 100-image stratified subset of the full 800-image dataset. While the subset was balanced across disease categories, results on the full dataset may differ. The width-stratified sensitivity metric depends on the accuracy of ground truth annotations at vessel boundaries, which varies across datasets. Finally, this study used five-fold cross-validation rather than official train/test splits, which limits direct numerical comparison with published SOTA results on DRIVE and CHASE_DB1. The study is intended as a controlled internal measurement rather than a benchmark comparison. Statistical tests are reported in Section 4.5. With only five folds, the minimum achievable two-sided p-value for the Wilcoxon signed-rank test is 0.0625, limiting the statistical power of between-condition comparisons.

# 6. Conclusion

This paper presented a controlled empirical study of the effect of input resolution on retinal vessel segmentation performance across five publicly available fundus datasets spanning a 6x range of native image widths. A fixed UNet architecture (1.9M parameters) was trained at five downsampling ratios per dataset with all other settings held identical, and performance was evaluated using both standard metrics and a width-stratified sensitivity metric that scores thin, medium, and thick vessel detection separately. The full numerical results across 25 conditions are reported in Table 3 and illustrated in Figures 1 and 2.

The central finding is that the effect of input resolution on thin vessel sensitivity depends on native image size. For high-resolution datasets (HRF, FIVES), downsampling consistently improves thin vessel detection by up to 9.4 percentage points, as it brings vessel structures into a range where the encoder can represent them as multi-pixel features. For low-to-mid resolution datasets (DRIVE, STARE, CHASE_DB1), native resolution already provides the best thin vessel sensitivity, and downsampling degrades performance by up to 15.8 percentage points. This dichotomy reflects the interaction between native vessel width and the receptive field capacity of the four-stage encoder.

Dice score does not capture these resolution effects. On HRF, Dice remained stable across conditions (0.8078 to 0.7725) while thin vessel sensitivity varied by 6.5 percentage points, demonstrating that high Dice can coexist with substantially degraded thin vessel detection. The practical implications of this study are threefold: (a) input resolution must be chosen relative to both native image size and encoder pooling depth; (b) Dice score alone is insufficient for evaluating retinal vessel segmentation across datasets with different native resolutions; (c) and the width-stratified sensitivity metric introduced here provides a reproducible tool for making these effects visible in future studies.

## Funding

The author declares that no funds, grants, or other support were received during the preparation of this manuscript. The work was conducted independently for academic and scientific purposes only, with no military, defense, or commercial application intended or involved.

## Competing Interests

The author has no relevant financial or non-financial interests to disclose.

## Ethics Approval

This is a computational study utilizing publicly available, fully anonymized datasets (DRIVE, STARE, CHASE_DB1, HRF, FIVES). No new human or animal data were collected. Therefore, ethical approval was not required.

## Consent to Participate / Publish

Not applicable, as the study relies exclusively on publicly available, de-identified datasets.

## Data Availability

All datasets used in this study are publicly available. DRIVE is available at https://drive.grand-challenge.org. STARE is available at https://cecas.clemson.edu/~ahoover/stare. CHASE_DB1 is available from the Kingston University data repository at https://researchinnovation.kingston.ac.uk/en/datasets/chasedb1-retinal-vessel-reference-dataset-4/. HRF is available at https://www5.cs.fau.de/research/data/fundus-images/. FIVES is available at https://figshare.com/articles/figure/FIVES_A_Fundus_Image_Dataset_for_AI-

based_Vessel_Segmentation/19688169/1. Source code, exact fold assignments, and trained model weights will be made publicly available on GitHub upon acceptance to support full reproducibility.